\documentclass{article}
\usepackage[export]{adjustbox}
\usepackage{subcaption}
\usepackage{graphicx} 

\usepackage[colorinlistoftodos]{todonotes}

\usepackage[]{main}
\usepackage{multirow}
\usepackage{arydshln}

\title{Privacy-Preserving Transformers: SwiftKey’s Differential Privacy Implementation}
\author{Abdelrahman Abouelenin, Mohamed Abdelrehim, Raffy Fahim, \\{\bf Amr Hendy and Mohamed Afify}
\\
{\bf Microsoft} \\}
  
\date{July 2024}

\begin{document}    

\maketitle

\begin{abstract}
    In this paper we train a transformer using differential privacy (DP) for language modeling in SwiftKey.
    We run multiple experiments to balance the trade-off between the model size and run-time speed and accuracy.
    We show that we get small and consistent gains in the next-word-prediction and accuracy with graceful increase in memory and speed compared to the production GRU. This is obtained by scaling down a GPT2 architecture to fit the required size and a two stage training process that builds a seed model on general data and DP finetunes it on typing data. The transformer is integrated using ONNX offering both flexibility and efficiency.
\end{abstract}

\section{Introduction}

Transformers provide state-of-the-art performance for many natural language processing (NLP) and language modeling tasks. Since their introduction for machine translation in \cite{vaswani2017} they showed impressive performance on multiple domains, mainly by an exponential increase in the number of parameters. We explore using transformers for context modeling in the SwiftKey keyboard. This requires deploying the model on-device and puts strict constraints on the model size. Also, transformers are usually trained on general data like web data, textbooks and conversations. The latter heavily mismatch the characteristics of typing data entered by users on their keyboards. Hence, using language models trained on general data results in poor typing quality. An obvious solution is to train models directly on typing data. This poses  serious privacy concerns and risks leaking personal data typed by the users\cite{carlini2019secret}. We approach this problem by first pre-training models on general data followed by using differential privacy (DP)\cite{dwork2006dp} to fine-tune the resulting models on user typing data. This requires carefully tuning the training recipe to mitigate any performance loss due to DP fine-tuning.

In WWDC last year Apple announced that the new versions of iOS and macOS come with a transformer language model that will give users predictive text recommendations as they type. While they didn't announce the model details, some attempts to reverse engineer it \cite{jackcook2023apple} \footnote[1]{\url{https://github.com/jackcook/predictive-spy}} reveal a GPT-2 like model with 6 decoder blocks, 512 hidden dimension and 34M parameters. The vocabulary is BPE-based of size around 15K sub-words. In contrast, we use our transformer for core typing and train models as small as 6MB after quantization. 

The rest of the paper is organized as follows. We first give a quick primer on language modeling in SwiftKey and differential privacy in Section \ref{primer}. The transformer model development is then given in Section \ref{baseline}. This includes the train and test data, model architecture, training recipe, fluency (decoder) integration followed by results and comparison to our recurrent (GRU) models. We mainly show that we can get performance as good as the current production GRUs with slightly larger transformers. Additional experiments to improve the performance along multiple dimensions including training hyper-parameters and DP recipe are provided in Section \ref{experiment}. We then show initial results for relative positional embeddings which help alleviate length mismatch in Section \ref{relative}.
We finally summarize our findings in Section \ref{summary}.


\section{Language Modeling and Differential Privacy Primer}
\label{primer}
This section is a quick primer on language modeling in SwiftKey and differential privacy (DP). This is not meant as an in-depth overview but rather provides context for the rest of the paper. 
\subsection{Language Modeling in SwiftKey}

Language models calculate the probability of a word given the previous $N$ words i.e. $p(w_{i}|w_{i-1}......w_{i-N})$. In Swiftkey, these are referred to as context probabilities and are used in the search, with other knowledge sources, to predict the best words given the user input. Initially, n-grams \cite{chen-goodman-1996-empirical} were used for language modeling. They provide excellent performance for short contexts but can't properly model longer sequences and don't generalize well to unseen events. To address these limitations, gated recurrent units (GRU) \cite{nosouhian2021gru} were later deployed to a subset of locales having sufficient data and were trained on user typing data. Due to privacy regulations, e.g. GDPR, the user data was limited to data from or before 2018. This results in poor modeling of recent data. Last year we used differential privacy (DP), more on this below, to train GRUs on more recent typing data and update GRUs for a number of locales.

Transformers \cite{vaswani2017} are currently showing state-of-the-art performance for language modeling, they are being adopted by competitive keyboards and can unlock different scenarios, like sentence completion and grammar correction, in addition to core typing. In spite of their attractiveness, transformers pose some challenges regarding real-time speed, memory and DP fine-tuning. We will address these challenges below to ship transformer for core typing as mentioned above. We will also benchmark the resulting transformers against production GRUs. 





\subsection{Differential Privacy}
Fine-tuning language models on domain data is a crucial step to achieve better in-domain results. It has been widely used for language modeling where a pre-trained model is built using a large amount of data and then fine-tuned on specific domain data to improve it's predictive capabilities on this domain e.g. \cite{zheng2024finetuning}.
Unfortunately, a problem could emerge here if the data contains any private information. In this case, the model becomes vulnerable to membership attack where the attacker can deduce if a specific data point was used to fine-tune the model.

In our case to safely fine-tune the model on user data, we need to guarantee that it is resilient to membership attack.
This is the main role of differential privacy (DP) in our training pipeline. Intuitively, DP is a randomized algorithm that can be applied to any algorithm to guarantee that any of its output is not heavily influenced by the presence of a single data point.

An algorithm $M$ is $(\epsilon,\delta)$ differentially private if for all \(X\) and \(X'\)  and for events \(S\) 

\[Pr[M(X)\in S]\le e^\epsilon Pr[M(X')\in S]+\delta\]

Where \(X\) and \(X'\) are neighboring datasets that differ only in a single data point.
One of the major advantages of DP is to quantify the privacy guarantees using \(\epsilon\) and \(\delta\).
The lower the value of \(\epsilon\) the tighter the privacy guarantees. Training DP neural networks is usually implemented using the DP SGD proposed in \cite{abadi2016}
and its variants.

The standard deviation of the noise added at each step is a function  $\epsilon$, $\delta$ and gradient clipping $C$.
For Gaussian mechanism, this is given by $C^2*\sqrt{2log(1.25/\delta)/}\epsilon$ \cite{abadi2016}.
It is also shown that privacy scales with the number of iteration $k$ as $q\epsilon\sqrt{k}$ where $q=L/N$ where $L$ is the batch size and $N$ the total size of training data. We will experiment with some of these parameters in Section \ref{experiment}.



\section{Model Development}
\label{baseline}
In this section we describe in detail the development of a transformer model and compare it to the current production GRU. This includes the train and test data, model architecture, training recipe, fluency (decoder) integration and results.

\subsection{Train and Test Data}
In this section we describe the train and test data used in our experiments. 






For training we use pre-training data for the seed model and typing data to obtain the final model. The pre-training data come from common crawl and twitter and consists of two sets:
\begin{itemize}
\item Pre-training I: A relatively small subset for fast turn-around that consists of 171.6M sentences and 2.9B tokens.
\item Pre-training II: The full set that consists of 479M sentences and 11.1B tokens.
\end{itemize}
 The fine-tuning data comes entirely from user typing data and also comprises two sets:
 \begin{itemize}
 \item Fine-tuning I: a relatively small set for fast turn around of experiments that has around 280M sentences, 934M tokens and 1.9M users.
 \item Fine-tuning II: a larger set that consists of around 628M sentences, 2.63B tokens and 6.8M users. 
 \end{itemize}
 For privacy, the typing data is anonymized by replacing all entities by a general placeholder. 

The test data also consists of two sets: 
\begin{itemize}
\item Snippets Set: The final models are tested on typing data. Full typing sessions which record the full user input are called snippets. For privacy, these snippets are limited to a maximum of 4 commits. These sessions allow calculating the typing accuracy \footnote{accuracy = (true positive + true negative)/  (true positive + true negative + false positive + false negative).}, or edit rate = 1 - accuracy,  as well as NWP. The Snippets data has size 6.6K sentences from January, 2024.

\item Typing Data Set: Snippets are generally short with an average length of 4. To better test for longer contexts, the typing data set comprises 50K sentences held out from the training data of lengths varying from 2 to 15 tokens with an average of around 7 tokens. We report NWP and simulated accuracy\footnote{Typing data is simulated using a keypress model since no typing data is available for this set.} for this set.  
\end{itemize}

\subsection{Vocabulary Construction}
Vocabulary selection is crucial for good typing accuracy.
We use a word-based vocabulary and will consider sub-word vocabulary in future work.

For vocabulary selection, we use a frequency-based approach. We combine the pre-training data and the weighted typing data\footnote{In initial experiments we found that a weight of 5 gives best results.} and then select the top N most frequent words. Currently, we use a vocabulary of size 20K words that gives a reasonably good trade-off between out-of-vocabulary (OOV) rate and model size. Table \ref{tab:oov-vocab-analysis} shows the OOV rate for different vocabulary sizes for en\_US evaluated on the BUS 50k test set that has ~19k unique words. It clearly shows that the OOV improves significantly by going from 10K to 20K vocabulary.

While a 20K vocabulary gives a good compromise between coverage and model size it still has significant OOV rate. 
One way to overcome this, that we will explore in future work, is to use a sub-word vocabulary.
Here, we use a back-off unigram language model to account for the unknown words i.e. to give non-zero probabilities to words outside the main vocabulary. The vocabulary of the unigram model is constructed from the most frequent 64K words in the data and not in the neural model vocabulary. The OOV rate after adding the unigram model is also shown in Table \ref{tab:oov-vocab-analysis}\footnote{While the OOV rate for the three vocabularies is still relatively high it is compensated by the user dynamic vocabulary but this is out of scope of this paper.}.



\begin{table}
    \centering
    \begin{tabular}{c|c}
        Vocab & OOV Rate \\
        \hline
        64k & 24.70\% \\
        + 64k Unigram & 23.59\% \\ \hdashline
        10k & 58.14\% \\
        + 64k Unigram & 24.72\% \\ \hdashline
        20k & 37.81\% \\
        + 64k Unigram & 23.92\% \\
    \end{tabular}
    \caption{OOV Analysis on different vocab sizes based on BUS 50k testset}
    \label{tab:oov-vocab-analysis}
\end{table}

\subsection{Model Architecture}
In this section we describe the model architecture used in this work. A transformer decoder architecture similar to GPT2 is employed but scaled down to satisfy on-device memory and speed requirements. We build a 4-layer model with 4 attention heads and a hidden dimension of 512.

\begin{figure}
    \centering
    \includegraphics[width=1\linewidth]{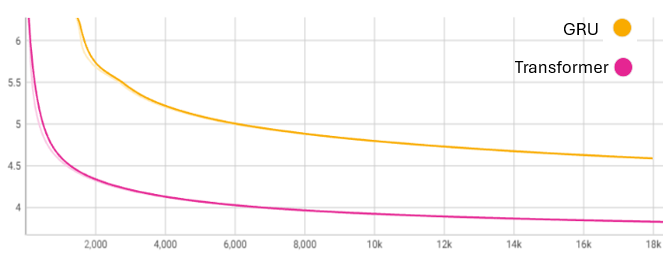}
    \caption{Pretraining evaluation loss of GRU and transformer model with same number of parameters. Eval set consist of 100k sentences taken from pretrained data}
    \label{Transformer vs GRU}
\end{figure}

The input embeddings take a substantial part of the total model size. Therefore, we limited the embedding dimension to 128. In initial experiments, we found that there is very small difference in performance between embedding size of 128 and 256. Finally, one byte quantization is applied to all the parameters. The total model size is around 6MB after quantization. 
To validate the selected architecture, we trained a transformer on the seed data and compared it to a similar size GRU model\footnote{The GRU has a 20K vocabulary, 512 units in the hidden layer.}. 
The evaluation loss of the two models during the training is shown in Figure \ref{Transformer vs GRU}. It is clear that the transformer has significantly lower evaluation loss for the same number of parameters. Based on this encouraging result, we finetuned the resulting transformer on typing data and compared it to the production model as will be discussed below.





\subsection{Training Recipe}
Our training recipe is divided into three main parts.
Data preparation and vocabulary creation, pretraining on general data and DP fine-tuning on reconstructed typing data. These steps are automated to eliminate manual intervention and facilitate training scaling.

\subsubsection{Data Preparation}
Data preparation is done using our internal tools which comprise input retrieval and vocabulary and quality filters. The retrieval part is straightforward and comprise fetching the data from the corresponding stream.
The vocabulary filters mainly remove profane and non-linguistic words using some predefined dictionaries while quality filters are high level filters to ensure quality of training data e.g. removing non-linguistic events or text in a different language.

\subsubsection{Model Training}
Model training, either pretraining or DP finetuning, is done using the DP\_transformer library \footnote{https://github.com/microsoft/dp-transformers/tree/main}. This library has a flexible architecture that support huggingface models and allows training transformers in DP and non-DP modes by integrating Opcaus \cite{opacus} with huggingface.

\begin{figure}
    \centering
    \includegraphics[width=1\linewidth]{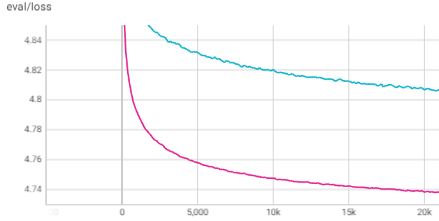}
    \caption{eval loss dp vs non dp}
    \label{dp_vs_non_dp}
\end{figure}

As mentioned above, we first train a seed model on general data then DP finetune it on typing data to create the final model. DP training has two important parameters that control the amount of privacy versus accuracy, namely $(\delta,\epsilon)$. In our initial experiments we fix $\delta$ at $10^{-8}$ and target $\epsilon$ at 14 i.e. it will increase throughout the training until it reaches the target value. This allows us to calculate the noise level that will be added to the gradient at each iteration. We finetuned our seed model for 28k updates with an effective batch size of 64K sentences and learning rate 1e-4. We also finetuned the same model without DP. The evolution of the loss for both DP and non-DP finetuning is shown in Figure \ref{dp_vs_non_dp}. We can observe some performance loss due to DP finetuning. This is the price to pay for increased privacy. We will run some experiments to mitigate the loss in Section \ref{experiment}.

\subsection{Fluency Integration}
In order to make the transformer model compatible with a generic framework, a crucial step involves converting it into the Open Neural Network Exchange (ONNX) format. This conversion facilitates seamless integration of the model into various frameworks and enhances its accessibility and usability across different platforms. The model's forward function is modified to return the probabilities for the next word prediction only instead of the whole sequence and a special handling for empty sequences is added.
While the current ONNX conversion lacks caching state handling, efforts are underway to address this, aiming to optimize the model's performance and speed.
Swiftkey keyboard is powered by an internal engine called "Fluency" where the context model (tranfsormer model) is a part of multiple search techniques contributing to its probability distribution for the next word prediction. In Fluency the ONNX model is loaded and used as part of inference.

\begin{figure}
    \centering
    \includegraphics[width=1\linewidth]{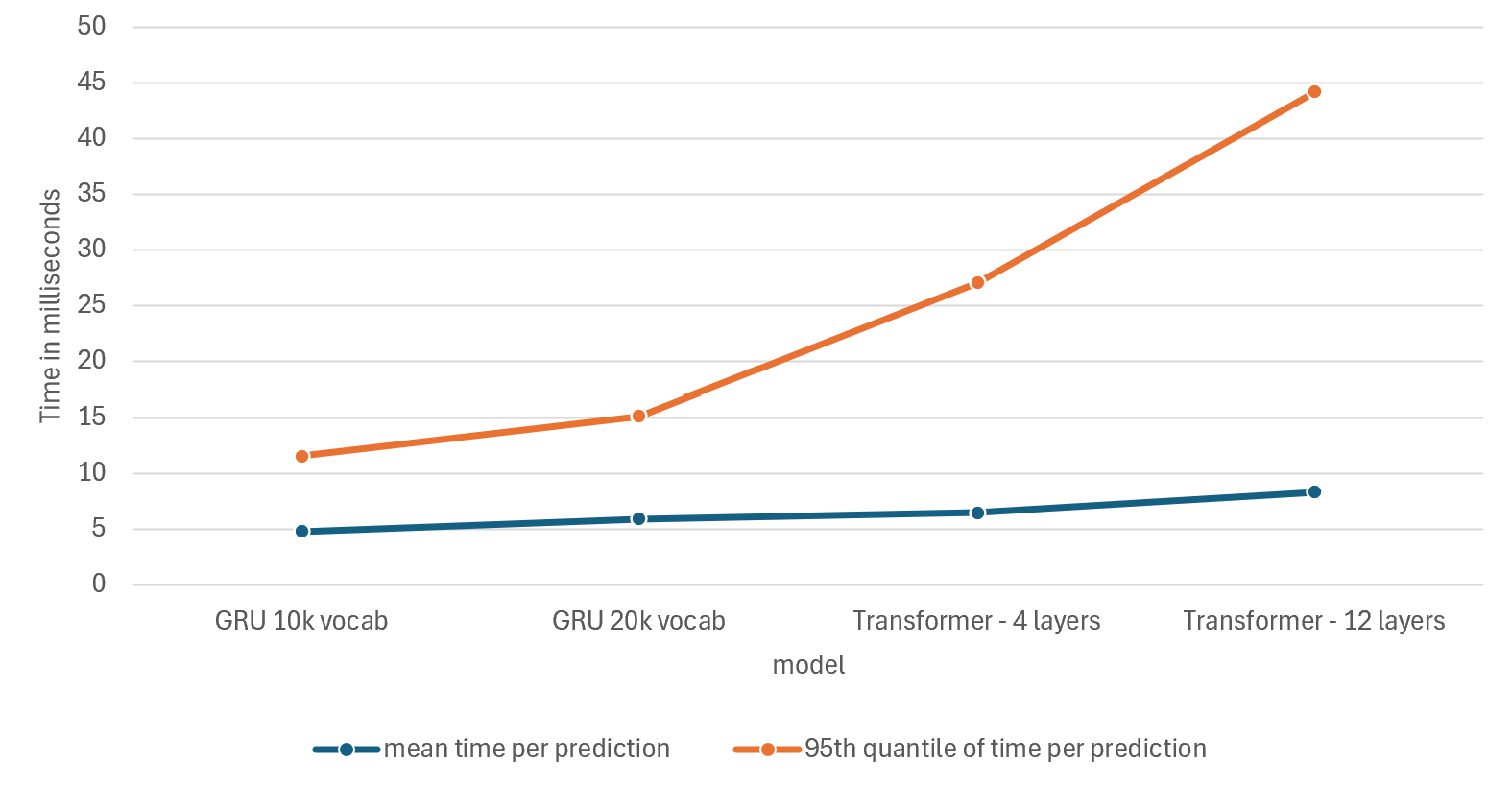}
    \caption{Inference Time}
    \label{fig:perf-test}
\end{figure}

\begin{figure}
    \centering
    \includegraphics[width=1\linewidth]{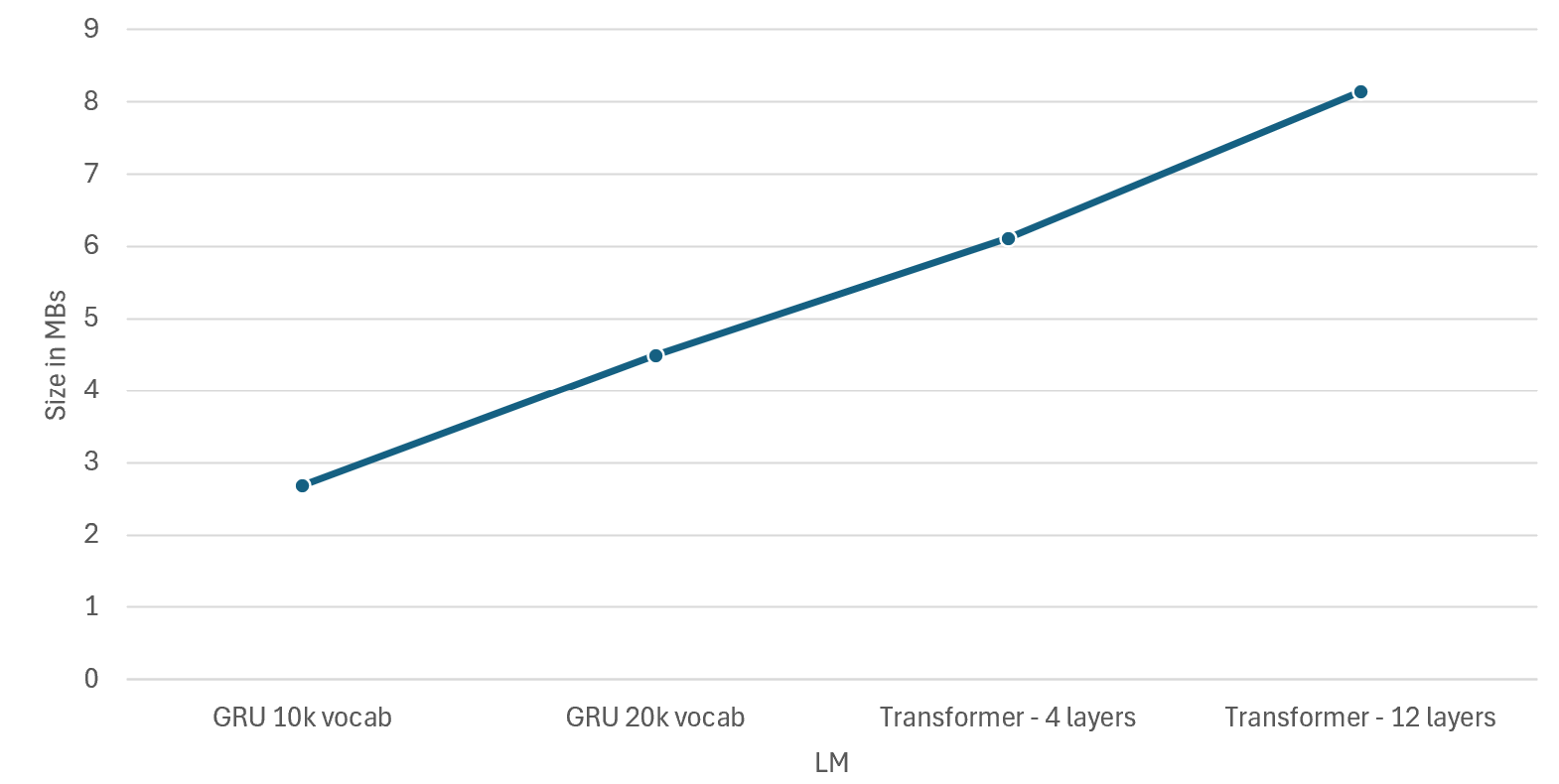}
    \caption{Different LM Sizes}
    \label{fig:lm-size-MBs}
\end{figure}

Performance tests are conducted to evaluate the speed of the transformer model and compare it to the production GRU for typing. A dataset composed of 630 typing samples is used. The sample is incrementally given as input to Fluency to predict the next word and each input is treated as an event where its time is calculated. Figure \ref{fig:perf-test} shows the mean and 95th quantile inference time in ms for the production GRU, a 20K vocabulary GRU, a 4-layer transformer and a 12-layer transformer. It can be observed that the 4-layer transformer gracefully increases the mean inference time while a deeper 12-layer transofrmer significantly increases it with respect to the GRU. The increase is more significant for the 95th quantile as longer sequences take more time due to the lack of caching for the transformers unlike for the GRU which caches hidden states for the sequence as we add a new word to it.
However, adding caching mechanism for the transformer in our engine is in progress. Also Figure \ref{fig:lm-size-MBs} shows the LM size for different transformer and GRU architectures. Most notably, for the same vocabulary size e.g. 20K the transfomer  increases the size by 20-25\% compared to the GRU.



\subsection{Results}



In previous sections we showed that we can DP-train a transformer model with graceful increase in memory and speed requirements compared to the production GRU. In this section we first present offline results varying the architecture, the training data size and DP vs non-DP finetuning. In the tables, $L$ and $A$ stand for the number of layers and attention heads resectively. The train data referred to 300M uses the smaller pretrain and finetuning sets while the 600M uses the larger sets.
Following this we show flight results for the converged 4L 4A model vs production GRU.

\subsubsection{Offline Results}

Here we show offline results of the model. This includes the accuracy and NWP on snippets data and 50K BUS data. We compare the model against the production GRU with 10K vocabulary and a 20K vocabulary GRU. 

Table \ref{tab:vocab-results} shows results for the GRU trained on the full data set and (4L,4A) transformer trained on 300M set for both 10K and 20K vocabularies. These models are trained using DP. For GRU, we notice a significant gain in NWP and, to a lesser extent, the accuracy for both test sets by increasing the vocabulary size. The improvement is less for the transformer. 
This can be explained by the fact that the GRU is trained on larger data and that DP might hurt the performance of transformer for a smaller number of users.  
Generally, the gains on the 50K BUS set is larger than the snippets due to larger average length.

The first four rows in Table \ref{tab:arch-data-results} show results with and without DP for different number of transformer layers on the smaller training set. We clearly see that DP leads to a clear gap and that deeper transformer results in significant gains at the expense of increased memory and speed. In the next two rows we train the 4-layer transformer on the larger training set we see nice gains especially for the DP case due to the increased number of users. Finally, the last row shows the same model trained for more iterations showing some additional gains.

\begin{table*}[!htbp]
\centering
\vspace{-0.8em}
\scalebox{0.9}{
\begin{tabular}{c c c | c c | c c}
\hline
\textbf{Model Arch} & \textbf{Train Data (Millions)} & \textbf{Vocab Size}  & \multicolumn{2}{c}{\textbf{Snippets}} & \multicolumn{2}{c}{\textbf{50k BUS}} \\ \hline
& &  & \textbf{NWP} & \textbf{Accuracy} & \textbf{NWP} & \textbf{Accuracy} \\

GRU & 600 & 10K & 16.20 & \textbf{90.90} & 26.74 & 81.20 \\
GRU & 600 & 20K & \textbf{17.32} & 90.68 & \textbf{28.59} & \textbf{81.69} \\ \hdashline
4L 4A & 300  & 10k & 16.20 & 90.08 & 26.76 & 80.99 \\
4L 4A & 300  & 20k & 16.31$^*$ & 90.59$^*$ & 27.44$^*$ & 81.47$^*$\\
\end{tabular}}
\caption{Different vocab results. Best scores across different systems are marked bold. * denotes the best results among Transformer systems.}
\label{tab:vocab-results}
\vspace{-0.5em}
\end{table*}

\begin{table*}[!htbp]
\centering
\scalebox{0.9}{
\begin{tabular}{c c c | c c | c c}
\hline
\textbf{Model Arch} & \textbf{DP} & \textbf{Train Data (Millions)} & \multicolumn{2}{c}{\textbf{Snippets}} & \multicolumn{2}{c}{\textbf{50k BUS}} \\ \hline
& & & \textbf{NWP} & \textbf{Accuracy} & \textbf{NWP} & \textbf{Accuracy} \\

4L 4A & N & 300 & 17.05 & 90.55 & 28.70 & 81.66\\
4L 4A & Y & 300 & 16.31 & 90.59 & 27.44 & 81.47\\ \hdashline[0.5pt/4pt]
12L 4A & N & 300 & \textbf{17.39} & 90.54 & \textbf{30.03} & 81.74\\
12L 4A & Y & 300 & 16.89 & 90.54 & 28.56 & 81.62\\ \hdashline
4L 4A & N & 600 & 17.27 & 90.82 & 28.64 & 81.66\\
4L 4A & Y & 600 & 17.17 & 90.82 & 28.46 & 81.82\\ \hdashline
4L 4A Converged & Y & 600 & 17.34 & 90.83 & 28.76 & 81.84\\ 
\end{tabular}}
\caption{Results for different training data setups and system arch on DP and non-DP training. Best scores across different systems are marked bold.}
\label{tab:arch-data-results}
\vspace{-0.5em}
\end{table*}


\subsubsection{Flight Results}

\begin{figure*}[h!]
  \centering         \includegraphics[width=0.8\textwidth]{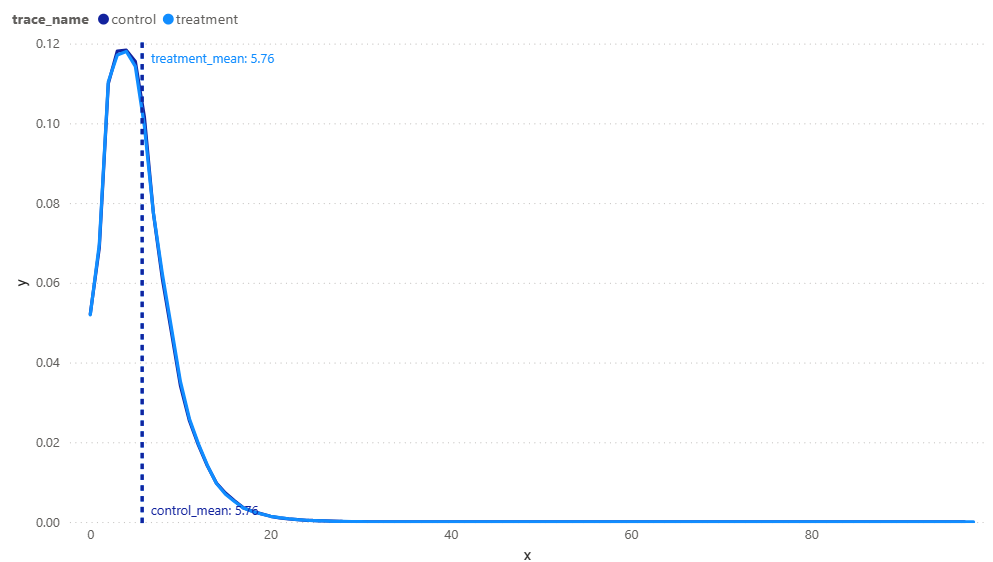}
        \vskip -0.1in
        \caption{\centering  Flight Results for GRU 10k vocab and Transformer 20k vocab trained on 600M training-set systems for two weeks.}
         \label{fig:flight_analysis_600m}
\end{figure*}

We run a market flight for the production GRU\footnote{First row in Table 2.} against the converged 4L transformer\footnote{Last row in Table 3.}.
We use Edit Rate (1-accuracy) as a metric. 
For each model, we randomly sample 1 million real user typing data from a 2-week period running flight. \autoref{fig:flight_analysis_600m} reports aggregated edit rate scores. We observe small difference in the aggregated edit rate. This is a bit in contrast to the clear gain in the offline results. The explanation is that the dynamic user model starts to kick in after certain period and override the improvements from the static language model. For this reason, we show the edit rate sliced by user age and notice some gains for the early ages in Table \ref{tab:flight-results-age-groups}.  


\begin{table*}
\centering
\vspace{-0.5em}
\scalebox{0.9}{
\begin{tabular}{c c c}
\hline
\textbf{Age Group} & 
\multicolumn{2}{c}{\textbf{Edit Rate}} \\

& \textbf{GRU 10k}  &
\textbf{Transformer 20k} \\ \hline

0-3 & 7.34 & \textbf{7.21} \\
4-7 & \textbf{6.68} & 6.70 \\
8-14 & \textbf{6.53} & 6.54 \\
15-90 & 6.25 & \textbf{6.23}  \\
90+ & 5.68 & 5.68 \\

\end{tabular}}
\caption{Flight results for different age groups}
\label{tab:flight-results-age-groups}
\vspace{-0.5em}
\end{table*}

\section{Additional Experiments}
\label{experiment}

Following the initial results in Section \ref{baseline} we conducted multiple experiments to optimize the performance. These include: a search over the training hyper-parameters mainly the batch size and learning rate as well as  
several aspects of DP finetuning like user sampling and gradient clipping. These experiments are conducted using the smaller training set and are discussed below.  




\subsection{Training Hyper-parameters}
\begin{figure}
    \centering
    \includegraphics[width=1\linewidth]{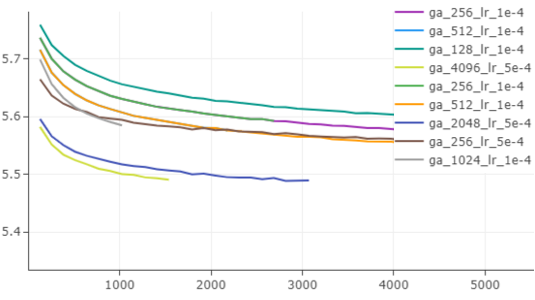}
    \caption{DP finetuning evaluation loss with different batch size and learning rate, batch size 2048 and 4096 are superior than smaller batch sizes.}
    \label{fig:eval loss}
\end{figure}

In this section we show the effect of varying the batch size (bsz) and learning rate (lr) when DP finetuning the model. Increasing the batch size generally improves the overall loss of the model but it significantly reduces the number of updates a model can achieve before reaching a target epsilon because decreasing the batch size reduces the privacy cost of each training step.
We can see in Figure \ref{fig:eval loss} that higher batch size reduces the evaluation loss but at the same time the training stops prematurely since it reached the target epsilon more quickly. Also larger batch sizes result in less noisy gradient estimates and hence we can safely increase learning rate for faster convergence. This is also clear in Figure \ref{fig:eval loss}.
We conclude that the best setting is an effective batch size (bsz)=32*512*4 = 64K and learning rate (lr)=1e-4.
However, higher batch size might be preferable for larger amount of data.

%
\subsection{Relaxed Author Sampling}
\begin{figure}[h] 
\begin{subfigure}{0.5\textwidth}
\includegraphics[width=0.9\linewidth, height=6cm]{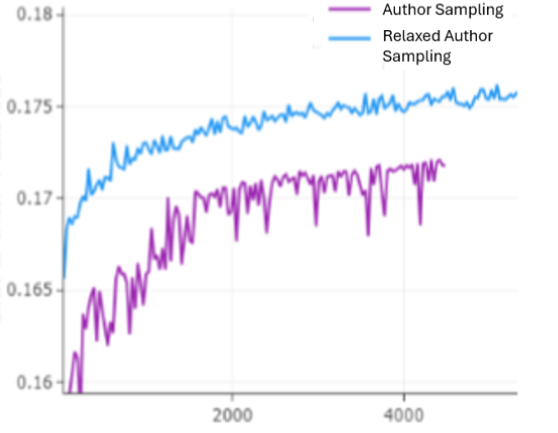} 
\caption{hit @ 3}
\label{fig:Sampling hit3}
\end{subfigure}
\begin{subfigure}{0.5\textwidth}
\includegraphics[width=0.9\linewidth, height=6cm]{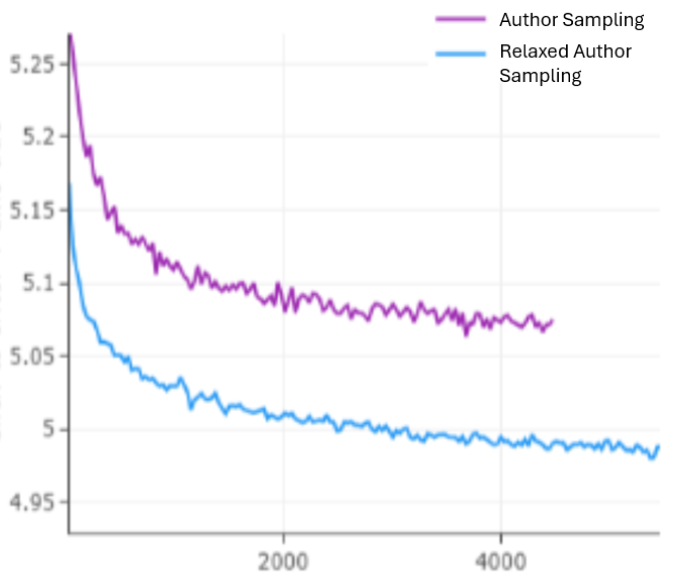}
\caption{Eval loss}
\label{fig:Sampling eval loss}
\end{subfigure}

\caption{comparison between author sampling and relaxed author sampling }
\label{fig:Sampling}
\end{figure}

We apply DP at user level mainly to make users indistinguishable in the resulting model.
One major problem here is that the number of data points per user are widely inconsistent. For example, some users have around 67k sentences while others have only 20 sentences. Therefore, sampling using the user distribution will result in batches that are dominated by a limited number of users and hence very poor results for most users.
To fix this issue, we sampled sentences while maintaining a uniform sampling probability across all sentences but imposing the constraint that any user occurs only once per batch. We refer to this as relaxed author sampling. Improving the sampling, reduced the evaluation loss and improved hit rate as shown in Figure \ref{fig:Sampling}.

\subsection{Gradient Clipping}
Gradient clipping is a crucial factor that affects the noise added and hence the training. This is due to the fact that the noise added at each step depends on the value of the gradient clipping. On one hand, higher clipping increases the sensitivity and therefore the noise and on the other hand a lower clipping value reduces the noise.
It would seem that we should always use the lowest gradient clipping value possible. However, reducing the clipping norm value degrades the overall model performance because it distorts the gradient.
Therefore, we conducted an experiment to see to which extent we should reduce the clipping value before we have a noticeable degradation in our model performance.

\begin{figure}
    \centering
    \includegraphics[width=1\linewidth]{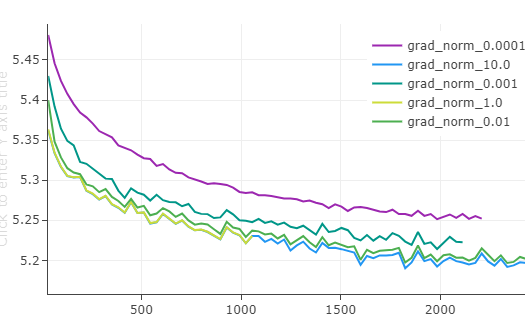}
    \caption{effect of Grad clipping}
    \label{fig:grad clipping}
\end{figure}

We can reduce the noise added by increasing the total amount of data but the clipping factor is a hyperparamter that needs tuning.
To this end, we removed the noise for all clipping norms and observed the effect of each on the training evaluation loss.
We can see in Figure \ref{fig:grad clipping} that decreasing the norm after 0.001 has a negative impact on the model training, while increasing the grad norm after 0.1 doesn't quite improve the training and only result in more noise being added.
Due to this trade off, we concluded that using a grad clipping norm of 0.01 is suitable for our case.




\section{Positional Encoding}
\label{relative}
\begin{figure}
    \centering
    \includegraphics[width=1\linewidth]{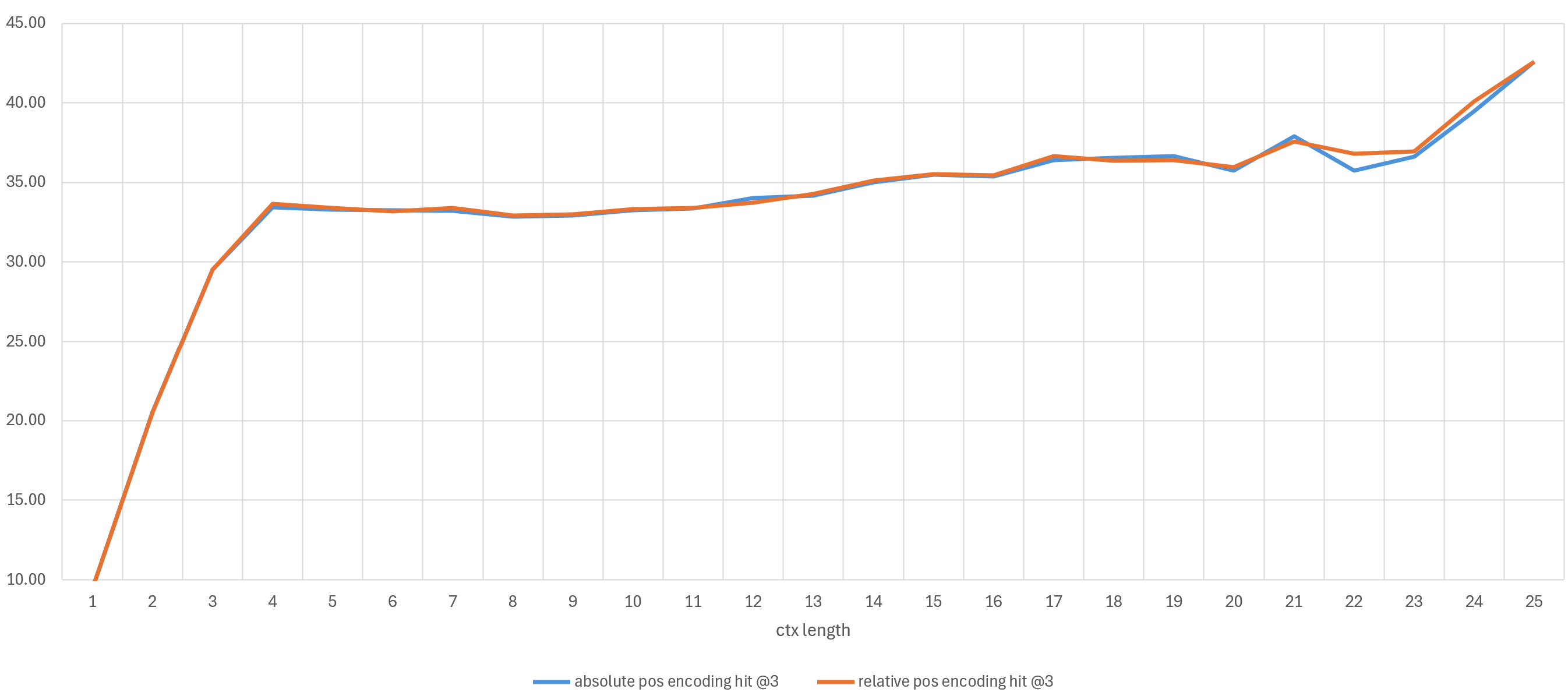}
    \caption{Hit @3 on different at different lengths on BU\&S testset.}
    \label{fig:length generalization}
\end{figure}

Our transformer model is trained using learnable absolute positional encoding. One drawback of this approach is that it cannot generalize beyond the maximum context length set in training since each position is represented as a row in an embedding table. Most of the recent transformer LLMs such as \cite{abdin2024phi} \cite{touvron2023llama} use some variation of relative position encoding since they show better length generalization \cite{li2023functional}. We did experiments where we replace the absolute positional encoding in our current architecture with FIRE \cite{li2023functional} positional encoding hoping to answer whether relative positional encoding provides benefits under context lengths seen in SwiftKey. Figure \ref{fig:length generalization} shows that there are no differences in results between relative and absolute positional encoding when testing with context lengths up to 25 words. We also noticed during our experiments that relative position encoding is robust to shifts in position ids. Table \ref{tab:position_ids_shifts} shows that the negative effect from shifting the position ids to start from a $0 \leq N < len(context) $ is much larger when using absolute position encoding compared to using relative position encoding. 
\begin{table}
    \centering
    \begin{tabular}{r|c|c}
        & abs & rel \\
        \hline
        no change & 30.04 & 29.13 \\
        shifted & 24.24 & 30.09 \\
    \end{tabular}
    \caption{The effect of shifting the position ids on hit @3 scores when using absolute position encoding vs using relative position encoding}
    \label{tab:position_ids_shifts}
\end{table}

\section{Summary}
\label{summary}
In this report we showed that we can DP-train a transformer for language modeling in SwiftKey. Compared to the production GRU we observe some small consistent gains in the next-word-prediction and accuracy with graceful increase in memory and speed requirements. This is obtained by scaling down a GPT2 architecture to fit the required size, a two stage training process that builds a seed model on general data and finetune it on typing data, and careful tuning of the training and privacy parameters. The transformer is shipped through ONNX conversion which is supported by the Fluency decoder and allows for a flexible framework for sequence-to-sequence context models. 

Areas of future work include training larger models and distilling them into the required size since we observed significant gains by increasing the model size, moving to subword models to improve the coverage and benefit from transformer longer context modeling and bridging the gap between the non-DP and DP training.

\section*{Acknowledgement}
We thank members of the Language and Intelligence team in London and Cairo for useful discussions and suggestions during this work.
\bibliography{anthology,main}
\bibliographystyle{acl_natbib}

\end{document}